\documentclass[letterpaper, 10 pt, journal, twoside]{IEEEtran}

\usepackage{verbatim}
\usepackage{amstext}
\usepackage{lipsum}
\usepackage{amsmath}
\usepackage{amsthm}
\usepackage{amssymb}
\usepackage{dsfont}
\usepackage{graphicx}
\usepackage{rotating}
\usepackage{dsfont}
\usepackage{csvsimple}
\usepackage{subcaption}
\usepackage{mathpartir}
\usepackage[normalem]{ulem}
\usepackage{booktabs}

\usepackage[shortlabels]{enumitem}
\usepackage{algorithm}
\usepackage{algpseudocode}
\usepackage{setspace}
\usepackage{gensymb}
\usepackage[dvipsnames]{xcolor}
\usepackage{soul}

\usepackage[hidelinks]{hyperref}

\usepackage{xcolor}
\newcommand{\jb}[1]{\textcolor{red}{[Joydeep: #1]}}

\usepackage{siunitx}

\newcommand{\algname}{\textsc{Plunder}}

\newcommand{\eg}{\textit{e.g.,}}

\newcommand{\ie}{\textit{i.e.,}}

\newcommand{\blue}[1]{{\color{blue} #1}}

\newcommand{\old}[1]{{\color{red}}}
\newcommand{\new}[1]{{\color{black} #1}}
\newcommand{\todo}[1]{{\color{orange}}}

\newcommand{\newfinal}[1]{{#1}}

\newcommand{\algcmt}[1]{\color{rule_label_color}{// \text{{#1}}} \color{black}}
  \definecolor{rule_label_color}{rgb}{0, 0.3, 0.0}

\DeclareMathOperator*{\argmax}{arg\,max}
\newcommand{\ha}[1]{\mathtt{#1}} 
\newcommand{\guard}[0]{\phi}

\newcommand{\asp}{\ensuremath{\pi}}

\newcommand{\stx}[1]{{\mathtt {#1}}}
\newcommand{\dns}{d_{\mathrm{stop}}}
\newcommand{\tabheader}[2]{\multicolumn{1}{#1}{#2}}
\newcommand{\ruleEq}[0]{:=\ \ }

\definecolor{my_cyan}{HTML}{559f79}
\definecolor{my_red}{HTML}{ce6262}
\definecolor{my_blue}{HTML}{567fca}
\definecolor{stx_color}{rgb}{0, 0, 0.8}

\newcommand{\ACC}{\textcolor{my_red}{\ha{ACC}}}
\newcommand{\CON}{\textcolor{my_cyan}{\ha{CON}}}
\newcommand{\DEC}{\textcolor{my_blue}{\ha{DEC}}}

\newcommand{\ALT}{\,\mid\,}

\begin{document}

\title{Programmatic Imitation Learning from Unlabeled and Noisy Demonstrations
}

\author{Jimmy Xin\textsuperscript{*}, Linus Zheng\textsuperscript{*}, Kia Rahmani, Jiayi Wei, Jarrett Holtz\textsuperscript{$\dagger$},  Isil Dillig, and Joydeep Biswas\\
The University of Texas at Austin \;\;\;\;\;\;\;\;\;\;\;\;\textsuperscript{$\dagger$}Robert Bosch LLC
\thanks{\thefootnote{*}These authors contributed equally to this work.}
}




\maketitle


\begin{abstract}
Imitation Learning (IL) is a promising paradigm for teaching robots to perform
novel tasks using demonstrations.
Most existing approaches for IL utilize neural networks (NN), however, 
these methods suffer from several well-known
limitations: they 1) require large amounts of training data, 2) are hard to
interpret, and 3) are hard to \old{repair }\new{refine} and adapt.
There is an emerging interest in \emph{Programmatic Imitation Learning} (PIL), which offers significant promise in addressing the above limitations. 
In PIL, the learned policy is represented in a programming language, making it amenable to interpretation and \old{repair }\new{adaptation to novel settings}. However, state-of-the-art PIL algorithms assume access to action labels and struggle to learn from noisy real-world demonstrations.
In this paper, we propose \algname, a
novel PIL algorithm that 
addresses these shortcomings by
synthesizing \emph{probabilistic} 
programmatic policies that are particularly well-suited for modeling the 
uncertainties inherent in real-world demonstrations. Our approach leverages an EM loop to 
simultaneously infer the missing action labels and the most likely probabilistic 
policy.
We benchmark \algname{} against several established IL techniques, and demonstrate its superiority across five challenging imitation learning tasks under noise. 
 \algname\ policies outperform the next-best baseline by $19\%$ and $17\%$ in matching the given demonstrations and successfully completing the tasks, respectively.

%

\end{abstract}

\IEEEpeerreviewmaketitle

\vspace{-2mm}
\section{Introduction}
\label{sec:intro}

\IEEEPARstart{I}{mitation} Learning (IL) is a popular approach for teaching robots how to perform
novel tasks using only human demonstrations, without the need to
specify a reward function or a system transition function~\cite{9927439}.
%
Most current IL approaches use neural networks to represent the learned policy,
mapping the agent's state to actions. While effective, these approaches have
several well-known limitations: they require large amounts of training data~\new{\cite{sunderhauf2018limits}};
 they are \new{opaque and} hard to interpret~\new{\cite{topin2019generation}}; and they are hard to \old{repair}\new{refine} and adapt to novel
settings\new{~\cite{chebotar2019closing}}.
%
Seeking to address these limitations, \textit{programmatic imitation learning}~(PIL) approaches synthesize \emph{programmatic policies} from demonstrations,
which are human-readable, require significantly fewer demonstrations, and are
amenable to refinement and adaptation~\cite{ldips,patton2023program,orfanos2023synthesizing}.



However, existing PIL methods require \textit{action labels} for  human 
demonstrations, which are often unavailable in real-world settings. Furthermore,
these methods assume that the demonstrations are \textit{nearly noise-free},
which is rarely the case in practice.
{In this paper, we introduce two key
insights to address the above challenges. First, given human demonstrations without
action labels, inferring the action labels is a \textit{latent variable
estimation} problem. Second, instead of synthesizing a deterministic policy,
synthesizing a \textit{probabilistic} policy allows us to model the
uncertainties inherent in real-world demonstrations~\cite{paraschos2018using}.}
Combining these insights, we introduce \algname, 
a new PIL algorithm that synthesizes \emph{probabilistic programmatic policies} from unlabeled and noisy demonstrations.

\begin{figure}[t]
    \centering
    \includegraphics[width=0.485\textwidth]{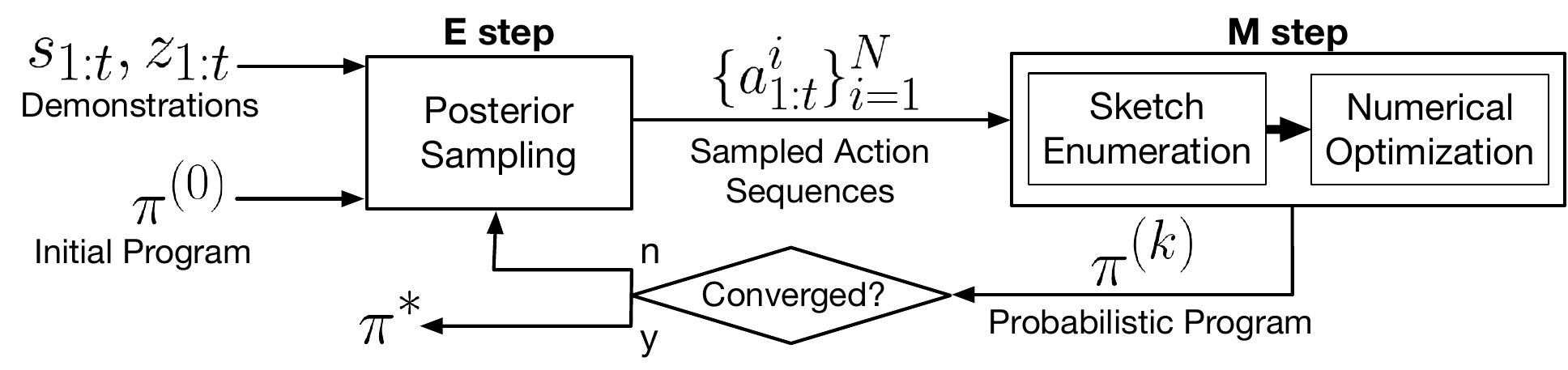}
    \caption{Overview of \algname{}}
    \label{fig:overview}
    \vspace{-5mm}
\end{figure}

\autoref{fig:overview} presents an overview of \algname{}.
The algorithm starts with an initial randomized policy ($\pi^{(0)}$)
and iteratively improves this policy while also improving its estimates of the action labels
using an Expectation-Maximization (EM) algorithm~\cite{em}.
In the E step, \algname{} samples posterior action label sequences by combining the current policy with the given demonstrations.
In the M step, it synthesizes a new policy that maximizes the likelihood of
producing actions that match the previously sampled action labels. To improve the scalability
of the M step, we also propose an incremental synthesis technique that narrows
the search on policies similar to the best policy found in the previous iteration.
This process is repeated until convergence, yielding jointly an optimal
policy $\pi^*$ and the most likely action labels for the demonstrations.

To evaluate our approach, we apply \algname{} to {five standard imitation learning tasks} and compare the results with multiple baselines, including \old{three}\new{four} state-of-the-art IL techniques. 
%
We empirically show that \algname\ synthesizes policies that conform to the demonstrations with  $95\%$ accuracy, which is $19\%$ higher than the next best baseline. Furthermore, \algname's policies are $17\%$ more successful than the closest baseline in completing the tasks.
\old{We also present case studies showing the advantages of \textsc{Plunder}, including resilience to noise, scalability to complex probabilistic programs, and interpretability to the end user.} 

\new{In summary, the key contribution of this paper is \algname{}, which, to the best of our knowledge, represents the first \emph{probabilistic PIL} approach specifically designed to synthesize programmatic policies from 
unlabeled
and noisy demonstrations. We validate our approach through extensive empirical evaluation on challenging IL benchmarks.}
%
Our implementation of the algorithm and the results of our empirical evaluation are available on the project website for \algname\ {(\url{https://amrl.cs.utexas.edu/plunder/})}.

\section{Related Work} 
\label{sec:related}
%

{\textbf{\textit{Imitation Learning}:}}
Imitation Learning (IL) provides a promising framework for learning autonomous agent behaviors from human demonstrations~\cite{9927439}. Unlike Reinforcement Learning (RL), IL does not require explicit reward signals. Although RL has been successful in addressing numerous complex tasks~\cite{sutton2018reinforcement},
IL techniques are better suited for many domains where formulating effective reward objectives proves extremely challenging~\cite{boularias2011relative}.


Within the broad umbrella of IL techniques, Behavior Cloning (BC) and Inverse Reinforcement Learning (IRL) have gained notable attention. BC focuses on establishing a function that directly associates observations with actions~\cite{daftry2017learning, bojarski2016end}\new{, \cite{chi2023diffusion, florence2022implicit}}.
 In contrast, IRL aims to decode the intrinsic reward structure from the given trajectories and subsequently leverages RL for policy inference~\cite{ziebart2008maximum}.
IRL typically demands more sophisticated algorithms and a substantial amount of training data to infer the expert's inherent motivations. Its primary objective is to \textit{enhance} the expert's performance or \textit{transfer} the acquired knowledge to related tasks~\cite{brown2019extrapolating}. 
In contrast, our approach in \algname{} aligns more closely with the BC setup, where a rapid and precise replication of expert behaviors is needed, and access to a complete simulation of the system is not available.

\old{The Generative Adversarial Imitation Learning (GAIL)  approach is a widely recognized IL technique that draws insight from both BC and IRL
GAIL concurrently trains a policy network and a discriminator network through an adversarial mechanism,
and has shown superior performance, particularly in large, high-dimensional environments.
%
However,  the neural policies trained by GAIL are not interpretable or repairable, and as such, it does not effectively address the PIL problem solved by \textsc{Plunder}. Furthermore, GAIL requires a complete simulation of the system, while \textsc{Plunder} necessitates only an observation model, which is often more accessible in practice.
%
}

\paragraph*{\new{\textbf{Programmatic Imitation Learning}}}
There is a growing interest in machine learning methodologies with enhanced \textit{interpretability}~\cite{molnar2020interpretable},
meaning that the learned models can be expressed using programmatic and human-readable structures like decision trees~\cite{orfanos2023synthesizing
} and finite-state machines~\cite{inala2019synthesizing,NEURIPS2021_d37124c4}.
%
For sequential decision-making tasks, programmatic policy inference has been studied under both RL and IL settings~\cite{topin2019generation}.
\new{Programmatic Imitation Learning (PIL) aims at learning such interpretable policy representations from a set of expert demonstrations~\cite{ldips, li2017infogail, perico2020learning}. }

\old{
In the RL setting, traditional gradient-driven methods such as DDPG and PPO have been employed to train a neural oracle, that is subsequently used for deriving a programmatic policy. 
Notably, NDPS
and PROPEL
utilize domain-specific context-free grammars to define the space of programmatic policies, and employ program synthesis to find the program within that space which best mimics the neural oracle.
Another interpretable RL technique is presented by Qiu et. al,
which concurrently learns program structures and parameters through a differentiable domain-specific language, obviating the initial neural oracle training.
Note that all the above strategies are designed for RL and rely on reward signals to learn a  policy. 
}

\old{
There are also multiple approaches for programmatic policy inference under the IL setting.
}

\new{
One of the early PIL approaches  was introduced in~\cite{tremblay2018synthetically}, which involves learning a human-readable plan from a single real-world demonstration. This plan is converted into a sequence of robotic actions suitable for broader applications. However, these plans provide only a basic description of the necessary state sequence for task completion. In contrast, the programs generated by \algname\ are more informative, explaining the specific conditions in the state space that trigger an action, and incorporate details on noise at decision boundaries.
}

A more recent PIL method is LDIPS~\cite{ldips},
that attempts to synthesize programmatic policies from human demonstrations. 
Similar to \algname{}, LDIPS uses a domain-specific language to enumerate program sketches, but it relies on a Satisfiability Modulo Theories (SMT) solver to find sketch completions. Moreover, LDIPS requires access to action labels and only synthesizes deterministic policies incapable of reasoning about noise in the demonstrations. This leads to lower performance,  as shown by our experiments. 

\newfinal{
Finally, PROLEX~\cite{patton2023program} is a recent PIL method for long-horizon tasks in complex environments. PROLEX synthesizes complex policies with nested loops and uses a Large Language Model (LLM) to leverage common-sense relationships between objects and their attributes for scalable synthesis. However, unlike \algname, which learns from low-level and noisy demonstrations, PROLEX is constrained to high-level and symbolic task demonstrations.
}

\paragraph*{\new{\textbf{Imitation from Observations}}}
\new{
\algname\ is related to a specific subset of IL research that focuses on Imitation from Observations (IfO)~\cite{torabi2018generative,liu2018imitation,torabi2019recent}. IfO aims to enable learning from existing demonstration resources, such as online videos of humans performing a wide range of tasks~\cite{zhou2018towards}. These resources provide information about the state of the environment, but they do not include the specific actions executed by the demonstrators.
For instance, a state-of-the-art IfO approach is GAIfO~\cite{torabi2018generative}, which aims to infer the state-transition cost function of an expert by concurrently training a policy network and a discriminator network through a generative adversarial mechanism~\cite{goodfellow2014generative}. 
However, the neural policies trained by GAIfO are opaque and lack interpretability, and as such, do not effectively address the PIL problem solved by \textsc{Plunder}.
\new{
Furthermore, GAIfO and similar neural network-based approaches often require significantly more training data than \algname{} to achieve accurate policies.}
}

\old{
A significant advantage of \textsc{Plunder} over many existing IL methodologies
is its fully \textit{offline} learning capability. Offline policy learning is often favored over its online counterparts because executing intermediate policies in simulation or the real world is usually slow, costly, and potentially hazardous.
%
Additionally, offline learning enables usage of historical datasets, thereby enhancing the method's applicability.

There are approaches that conduct IL in a fully offline manner. For instance, ORIL
adapts the GAIL framework for offline settings and learns a policy using a collection of logged experiences and expert demonstrations. 
ORIL first learns a reward model by contrasting the expert demonstrations and the logged experiences, and uses it to learn a policy via offline RL.
%
However, ORIL demands hundreds of expert demonstrations and many thousands of logged experiences, while also training an opaque neural policy. In comparison, \textsc{Plunder} needs fewer than a dozen trajectories and produces a fully interpretable policy.
}

\paragraph*{\textbf{Expectation Maximization Techniques}}
The Expectation Maximization (EM) algorithm is a commonly used technique in situations where both the generative model and its hidden states are unknown~\cite{em}. Recently, EM has been applied in various robotics domains\new{~\cite{blessing2023information}}. For example, in \cite{steady}, the authors propose using an EM loop to simultaneously learn a dynamics model and estimate the robot's state trajectories. 
In \cite{Inala2020Synthesizing}, an internal EM loop was employed to optimize a learned policy. However, this method is focused on RL and is not well-suited for handling noise. 

\vspace{-0.25em}

\section{Problem Formulation}

Given a set of demonstrations of a task, the PIL problem addressed in this paper is to infer a probabilistic \emph{Action Selection Policy (ASP)} that is maximally consistent with the given demonstrations and that captures the  behavior intended by the demonstrator. 

In particular, we consider learning policies over a continuous state space $S$
(\eg{} $\mathds{SE}(2)$ for a ground mobile robot) 
and a discrete action space $A$, where the actions are abstracted as \emph{skills}~\cite{7139412, SUTTON1999181}. For instance, an autonomous vehicle may have skills such as accelerate ($\ACC$), decelerate ($\DEC$), and maintain constant velocity ($\CON$). A probabilistic ASP, $\asp$, is a probabilistic program that, given the agent's current state ($s_i\in S$) and its current action ($a_i\in A$), defines the probability of taking the next action $a_{i+1} \!\sim\!\asp(s_i, a_i)$. 

The given demonstrations consist of an agent's \emph{observations} $z_{1:t}$ (\eg{} the vehicle's acceleration input), and the corresponding \textit{state} trajectories, $s_{1:t}$ (\eg{} the vehicle's velocity), but do not include any action labels. 
Our objective is to infer a \emph{maximum a posteriori (MAP)} estimate  of an ASP, as $\small 
\asp^* = \argmax\limits_{\asp} P(z_{1:t}, \asp|s_{1:t})$, which can be factorized as $\small \pi^*= \argmax\limits_{\asp} P(z_{1:t}|s_{1:t}, \asp) P(\asp)$.
%
%
%
%
We next introduce action labels, $a_{1:t}$, as marginalized latent variables which should be  estimated jointly with the ASP: 
\begin{equation} \small
   \asp^* = \argmax_{\asp} \sum_{a_{1:t}} P(z_{1:t} | a_{1:t}, s_{1:t}) P(a_{1:t}|s_{1:t}, \asp) P(\asp)\ \label{eq:objective}
\end{equation}
In the above formula $P(z_{1:t} | a_{1:t}, s_{1:t})$ is the \emph{observation model} 
that defines the likelihood of an observation trajectory given the agent's actions and states, and $P(a_{1:t}|s_{1:t}, \asp)$ is defined by the policy $\asp$. 
\new{Function $P(\cdot)$ denotes a prior distribution on ASPs. A specific instance of $P(\cdot)$ will be introduced in  \autoref{sec:alg_overview}, which aims to reduce overfitting to the data.}
We also make the Markov assumptions on ASPs and the observation   model,~\ie{} $\small
    P(a_{1:t}|s_{1:t},\asp) = \prod_{i=1}^t P(a_i|a_{i-1}, s_i, \asp),
    \label{eq:transition-prob}$ 
    and 
    $\small
    P(z_{1:t}|a_{1:t},s_{1:t}) = \prod_{i=1}^t P(z_i|a_i, s_i) 
    \label{eq:motion-model}$,
yielding our final objective: 
\begin{equation} \small
  \asp^*= \argmax_{\asp} \sum_{a_{1:t}} \prod_{i=1}^t P(z_i|a_i, s_i)\, P(a_i|a_{i-1}, s_i, \asp) \,P(\asp).\ \label{eq:final-objective}
\end{equation}

There are two major challenges in solving \eqref{eq:final-objective}. First, directly computing the sum over all action trajectories is computationally intractable. Second, the search space for probabilistic policies is prohibitively large, making any na\"ive search within this space unscalable.
In \autoref{sec:alg_overview}, we explain our approach to overcome both of these challenges in a tractable manner.

\subsection{Example: Stop-Sign}
\label{subsec:stop_sign}
\old{Consider an autonomous vehicle that is required to drive on a straight road and make appropriate stops at every stop sign.}
\new{We now illustrate our approach using the example of an autonomous vehicle tasked with driving along a straight road and making appropriate stops at each stop sign.}
Defining a reward function or an objective function that accurately captures the desired behaviors for executing this task is challenging.
%
Instead, we aim to learn an ASP from expert demonstrations of the task using imitation learning.

\autoref{fig:demos} presents 10 demonstrations for this task. Each demonstration consists of the vehicle's velocity and acceleration trajectories 
between two consecutive stop signs. 
%
%
 \old{Observe the variations in different executions of this maneuver.}
 \new{There are noticeable variations in these demonstrations.}
 For instance, some demonstrations prefer a more gradual acceleration and deceleration for a comfortable ride. In contrast, others showcase harder acceleration and deceleration.

As we will discuss in the subsequent sections, the \algname{} algorithm is capable of reasoning about variations and noise in demonstrations and can synthesize a probabilistic ASP that  captures the  behavior intended by the demonstrations collectively.

\def\scalefactor{0.22}
\begin{figure}[t]
\centering
  \begin{subfigure}[b]{\scalefactor\textwidth}
  \centering
   \hspace*{-0.4cm} \includegraphics[width=1.16\linewidth]{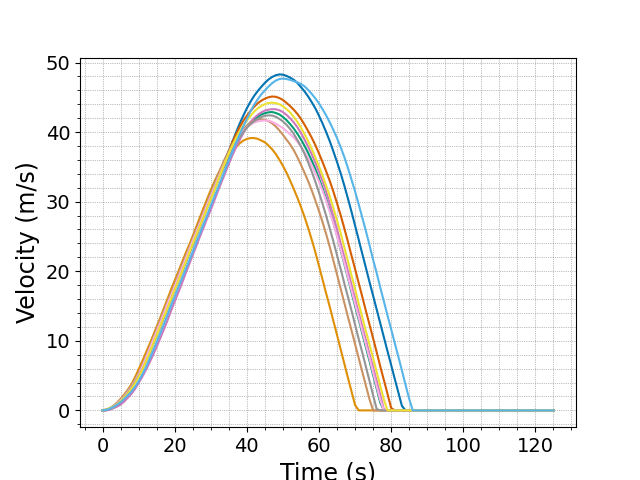}
    \caption{velocity/time}
    \label{subfig:demo_vel}
  \end{subfigure}
  ~
  \begin{subfigure}[b]{\scalefactor\textwidth}
  \centering
     \hspace*{-0.4cm}  \includegraphics[width=1.16\textwidth]{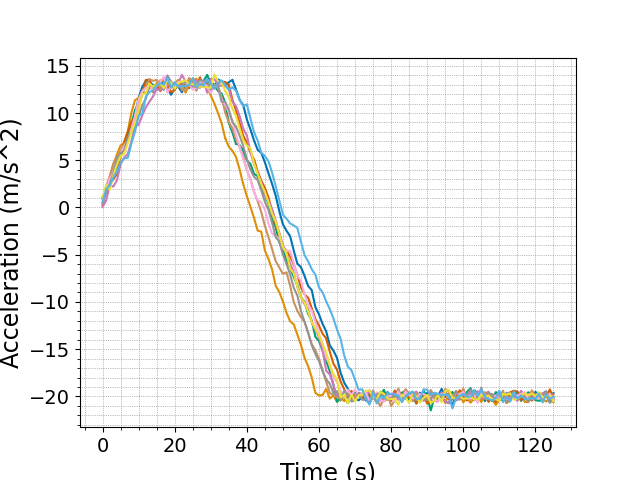}
    \caption{acceleration/time}
    \label{subfig:demo_acc}
  \end{subfigure}
  \caption{
  Demonstration trajectories for the Stop Sign task. The acceleration value of this particular vehicle 
  cannot exceed $a_{\max}\approx13m/s^2$ or drop below $a_{\min}\approx-20m/s^2$. 
  } \label{fig:demos}
  \vspace{-4mm}
\end{figure}

\begin{algorithm}[h]
\footnotesize
    \caption{The \algname{} Algorithm} \label{alg:overview}
     \textbf{Inputs:}  state trajectories: $\bar{S}$,\  observation trajectories: $\bar{Z}$,\ observation model: $O$,\ 
     initial policy: $\asp^{(0)}$,\ convergence threshold: $\gamma$
     \\
     \textbf{Output:} probabilistic action selection policy: $\asp^*$
    \begin{algorithmic}[1]\onehalfspacing
    \State $k=0$
    \While{$\mathtt{likelihood}(\bar{S}, \bar{Z}, O, \pi^{(k)})\leq \gamma$}
        \State $\{a_{1:t}^i\}_{i=1}^N := \mathtt{runPF}(\bar{S}, \bar{Z}, O, \pi^{(k)})$ 
        \ \ \algcmt{\texttt{E step}}
        \State $\pi^{(k+1)} := \mathtt{synthesize}(\pi^{(k)},  \bar{S}, \{a_{1:t}^i\}_{i=1}^N)$ 
        \ \ \algcmt{\texttt{M step}}
        \State $k = k + 1$
    \EndWhile
    \State \Return $\pi^{(k)}$
\end{algorithmic}
\end{algorithm}

\section{The \algname{} Algorithm}
\label{sec:alg_overview}
%


Algorithm~\ref{alg:overview} provides an overview of our Expectation-Maximization (EM) approach, which is designed to tractably approximate the MAP estimate in equation~(\ref{eq:final-objective}).
The inputs to the algorithm are a set of $D$ demonstrations consisting of state trajectories ($\bar{S}:=\{s^i_{1:t}\}_{i=1}^D$) and observation trajectories ($\bar{Z}:=\{z^i_{1:t}\}_{i=1}^D$), the observation model $O$, the initial policy $\asp^{(0)}$,
and a threshold, $\gamma$, for determining whether the synthesized policy has converged. 
%

The algorithm alternates between the E step (line 3) and the M step (line 4). 
In the E step, a posterior sampling of \(N\) action label trajectories is performed using the current candidate ASP \(\asp^{(k)}\) and the given demonstration trajectories. These sampled action trajectories are then used in the M step to synthesize a policy that is maximally consistent with those action trajectories. 
%
%
After each iteration of the E and M steps, the likelihood of the synthesized policy's prediction on the given trajectories is measured. Once the desired likelihood is reached (\ie{} the policy has converged), the algorithm outputs the latest synthesized policy as its final result, \(\asp^*\).

In the remainder of this section, we introduce the syntax of probabilistic ASPs synthesized in \algname{}, and present further details about the E and M steps of Algorithm~\autoref{alg:overview}.

%


\subsection{Probabilistic ASPs}

\def\scalefactor{0.4}
\begin{figure}[b]
  \vspace{-5mm}
 \centering
\footnotesize
\begin{mathpar}
\begin{array}{rll} 
\text{\scriptsize (Feature)\!\!\!} &
f & \!\!\!\!\ruleEq \!\! y_t \ALT c \ALT  g(f_1,\dots, f_n) 
\\
\new{\text{\scriptsize (Prob. Dist.)\!\!\!}} &
\new{\psi} & \new{\!\!\!\!\ruleEq \!\!r \ALT \stx{lgs}(f,x_0,k)} 
\\
\new{\text{\scriptsize (Guard)}\!\!\!}& 
\new{\guard_{}}\!\!\!\!\!\! & \new{\!\!\!\!\ruleEq \!\!\stx{flp}(\psi) \ALT \guard_1\wedge \guard_2 \ALT \guard_1\vee\guard_2} 
\\
\text{\new{\scriptsize (Trans. Seq.)\!\!\!}} &
\new{\tau} & \new{\!\!\!\!\ruleEq \!\!\stx{if}\; (\phi\; \stx{and}\; a_{t-1}\!=\!A) \;\stx{then}\; A'\;;\tau \ALT 
\stx{skip}} 
\\
{
 \text{\scriptsize {(ASP)}\!\!\!}} &
\pi(y_t, a_{t-1}) &{\!\!\!\!\ruleEq\!\! \tau}  
\\


\end{array}
  \end{mathpar}
  \caption{\new{Grammar of ASPs. Here, $y_t,a_{t-1}$ are inputs representing the current state and   previous action, $c$ is a constant, $A$ is an action, and $g$ is a built-in  ($+, \times$ etc) or domain-specific feature extraction function (e.g., {\tt timeToStop}).}}
  \label{syntax}
\end{figure}

\old{
In our setting, the set of agent's actions, A, is both discrete and finite. Therefore, at each time step, the policy must decide on the next action to take from this set. This decision is  conditioned on the agent's last action and state; thus,  the logic of policy $\asp$ can be defined using a set of binary transition conditions $\Phi:=\{\phi_{a,a'}\;|\; a,a'\in A\}$, where each $\phi_{a,a'}$ is evaluated on the agent's state.  
If the agent takes action $a$ at time $t$ and $\phi_{a,a'}$ holds true on the current state, then the agent will take action $a'$ at time $t+1$.  The policy also maintains an order among transition conditions and uses it to break ties when multiple conditions are true. }

\old{Since we want to synthesize probabilistic policies, each transition condition $\phi_{a,a'}(\cdot)$ 
is defined in a probabilistic Domain-Specific Language (pDSL), where Boolean expressions in each state are evaluated to 
$\stx{true}$ with a certain probability.
Figure~3 presents the pDSL used in Plunder. The numerical expressions in this pDSL are domain-specific and capture the physical properties of the system. 
}

\new{
\autoref{syntax}  presents a probabilistic Domain-Specific Language (pDSL)~\cite{probabilistic-robotics} used to compose  ASPs in \algname{}. A policy $\pi$ in this DSL takes as input the current state $y_t$ and previous action $a_{t-1}$ and returns a new action $a'$. 
A policy is a sequence of conditionals 
$\stx{if} (\phi \;\stx{and}\; a_{t-1} = A) \ \stx{then} \ A’$, guarding transitions from previous action $A$ to new action $A'$ based on the predicate $\phi$ that is evaluated on the current state $y$.
For brevity, we use the notation $\phi_{A, A'}$ to denote that $\phi$ is the guard associated with transition from action $A$ to action $A'$ with the understanding that it would comprise the statement $\stx{if} (\phi_{A,A'} \;\stx{and}\; a_{t-1}\!=\!A) \ \stx{then}\  A’$. Thus an ASP can be alternatively viewed as an ordered list of guards $\phi_{A_0, A_0'}, \ldots, \phi_{A_n, A_n'}$.


%
%
}

\new{Guards in this DSL are boolean combinations of probabilistic predicates, parameterized over policy parameters $r, x_0, k$ to be synthesized. Atomic predicates are of the form $\stx{flp}(\psi)$, which evaluates to true with probability $\psi$ and false with probability $1-\psi$. In the simplest case, $\psi$ is  a constant $r \in [0,1]$. In most cases, however, $\psi$ is a logistic function $\stx{lgs}(f, x_0, k)$, such that $\stx{flp}(\stx{lgs}(f, x_0, k))$ smoothly approximates the inequality $f \leq x_0$ using the logistic function parameter $k$ that controls the sharpness of the transition. Note that $f$ is  either the current state $y_t$ or a feature extracted from the current state using a built-in function, such as {\tt timeToStp} in the Stop-Sign domain. As in other programmatic policy synthesis algorithms~\cite{ldips,patton2023program,verma2018programmatically}, the feature extraction functions as well as the space of possible actions are domain-dependent.}


\old{Transition predicates are implemented using the probabilistic coin-flip function, $\stx{flp}(p)$, which evaluates to $\stx{true}$ with probability $p$ and evaluates to $\stx{false}$ otherwise. Within this coin-flip function, the S-shaped logistic function, $\stx{lgs}$, is used to  approximate numerical inequalities. For instance, the deterministic inequality $e>x_0$ can be transformed into a probabilistic inequality $\stx{flp}(\stx{lgs}(f,x_0,k))$. Here, $\stx{lgs}(f,x_0,k)$ maps the difference between $f$ and $x_0$ to a value in the range $(0,1)$ in a smooth and continuous fashion. The parameter $k$ controls the growth rate of the logistic curve. A larger value of $k$ indicates sharper growth, leading to less uncertainty in the probabilistic inequality. }

\old{Lastly, the function 
$\stx{randSwitch(r)}$ dictates that the policy retains the current action with probability $1-r$ or transitions to another randomly selected action with probability $r$. 
This function is stochastic, however, it embodies the \emph{``action inertia''} heuristic, suggesting that useful policies avoid switching between actions too frequently.}

\old{The numerical expressions and the set of action labels in this pDSL are domain-specific and are specialized for the Stop-Sign example. }




\subsection{Expectation (E) Step}
During the E step (line 3) of Algorithm~\ref{alg:overview}, 
the current estimate of the ASP, $\asp^{(k)}$, is used to sample $N$ plausible action sequences from the posterior distribution $\{a^i_{1:t}\}_{i=1}^N \sim P(.\,|\,s_{1:t}, z_{1:t}, \asp^{(k)})$. 
This posterior sampling problem is particularly amenable to inference using Monte-Carlo estimation algorithms,
and we use a \textit{particle filter}~\cite{doucet2009tutorial} to sample a policy's action label sequences.

The function \(\mathtt{runPF}(\bar{S}, \bar{Z}, O, \pi^{(k)})\) implements the particle filter as follows. For each state trajectory \(s_{1:t} \in \bar{S}\), the function utilizes \(\pi^{(k)}\) to sample action labels forward in time. Next, it employs the observation model \(O\) to re-weight and re-sample the particles. This ensures that sequences more aligned with the provided observations have a higher likelihood of duplication and representation in the final sample set. After completing this procedure for all provided demonstrations, the final particle set represents \(\{a_{1:t}^i\}_{i=1}^N\), \ie{} \(N\) plausible action label sequences given the current policy and the demonstration traces.

\subsection{Maximization (M) Step}
During the M Step (line 4) of Algorithm~\ref{alg:overview}, a new ASP is synthesized using the function $\mathtt{synthesize}(.)$. This function, given the state trajectories $\bar{S}$ and the sampled action sequences from the previous E step $\{a_{1:t}^i\}_{i=1}^N$, aims to solve the following objective:
\begin{equation} \small \label{eq:optimal-synthesis-}
    \asp^{(k+1)} = \argmax_{\asp\in \mathds{N}(\asp^{(k)})} \;
     \sum_{i=1}^N \log P(a^i_{1:t} \mid s_{1:t},\asp) + \log P(\asp)
\end{equation}
The above formula indicates that the synthesized policy should be as consistent as possible with the sampled action sequences and should have a high probability of occurrence based on the ASP prior. 
We define the ASP prior as 
$\log P(\asp) = -\lambda \cdot \mathrm{size}(\asp)$, 
where $\lambda\geq 0$ is a \new{regularization constant to promote smaller sized ASPs~\cite{gulwani2017program}.} 
\old{hyperparameter. This prior disincentivizes
synthesizing structurally complex programs such as programs with case-based reasoning that over-fit to inputs.}
\new{
Larger values of 
$\lambda$ discourage the synthesis of overly complex programs that overfit to the demonstrations.
}


The function $\mathtt{synthesize}(.)$ implements a bottom-up inductive synthesizer~\cite{synthesis-survey}, where the production rules described in \autoref{syntax} are used to enumerate a set of policy structures, or \emph{sketches}. 
The enumeration begins with the set of numerical {{features}} -- consisting of constants, variables, and function applications -- which are used to construct probabilistic threshold sketches with two unknown parameters. 
The threshold sketches
 are combined using disjunctions and conjunctions to construct the final set of policy-level sketches. 


 

 Each policy sketch can be converted into a complete policy by determining the unknown real-valued parameters in the probabilistic thresholds. 
 Our goal is to identify the parameters that optimize the objective given by equation (\ref{eq:optimal-synthesis-}).
%
%
%
To solve this optimization problem, we employ the Line Search Gradient Descent (L-BFGS) algorithm~\cite{liu1989limited}, selecting the top result out of four random initial assignments. After completing all sketches, the $\mathtt{synthesize}(.)$ function returns the optimal policy according to (\ref{eq:optimal-synthesis-}) as its final output.

%

While the above inductive synthesis approach is guaranteed to find the optimal solution within its bounded search space, enumerating programs this way rapidly becomes intractable. This is especially problematic in our setup, where the synthesis is integrated within the EM loop, which might need many iterations to converge.
To address this challenge, 
we restrict the search space for $\asp^{(k+1)}$ to $\mathds{N}(\asp^{(k)})$,  the \emph{syntactic \newfinal{neighborhood }}of $\asp^{(k)}$. 
To enumerate the syntactic neighborhood of a policy, we mutate its abstract syntax tree to derive a set of new expressions. In particular, we apply the following types of mutations:
\begin{enumerate}
\item Add a new threshold predicate with a random feature.
\item Simplify the policy by removing an existing predicate.
\item Swap conjunctions with disjunctions, and vice-versa.
\item Augment numerical features by applying random functions with random parameters. 
\item Simplify features by removing function applications.
\end{enumerate}
Each syntactic \newfinal{neighborhood}, in addition to the above set of mutated expressions, contains a set of atomic features. This allows the synthesizer to \textit{``reset''} to simpler policies if~needed. 

To further improve the scalability of the $\mathtt{synthesize}(.)$ function, we also prune the search space using a type system based on physical dimensional constraints, as previously done in~\cite{ldips}. This type system tracks the physical dimensions of expressions and discards those that are inconsistent.

The above two pruning methodologies significantly reduce the policy search space, allowing the $\mathtt{synthesize}(.)$ function to be tractably used within our EM framework.

\subsection{Example: Stop-Sign}

    \def\leftscalefactor{0.44}
    \def\rightscalefactor{0.44}
    \begin{figure}[t]
        \centering
          \begin{minipage}[c]{\rightscalefactor\textwidth}
        \footnotesize
 \hspace{4.68mm}\includegraphics[width=0.92\textwidth]{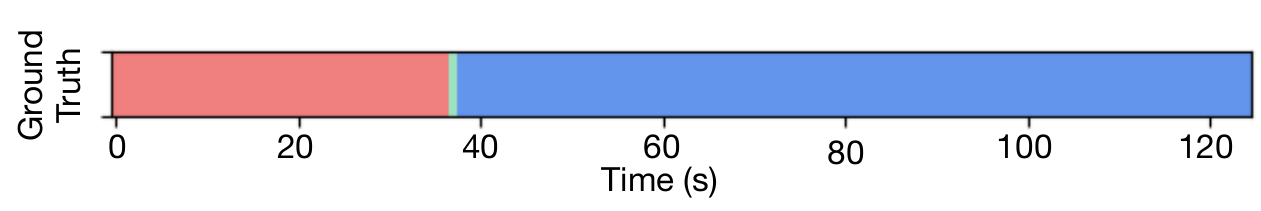}
        \end{minipage}
        \vspace{-2mm}
        \\ \centering
       \begin{minipage}[c]{\rightscalefactor\textwidth}
        \footnotesize
\begin{flalign*}
&\pi^{(0)}
    \begin{cases}
    \includegraphics[width=0.95\textwidth]{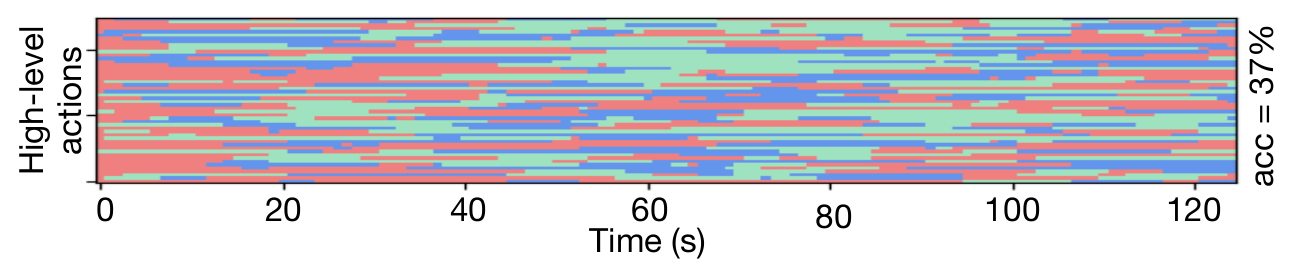}\\[-2mm]
       \;\;\;\;\;\;\guard_{{\ACC},{\CON}}\!:=\! \stx{flp}(0.1)   \\
         \;\;\;\;\;\;\guard_{{\CON},{\DEC}}\!:=\!\stx{flp}(0.1) 
    \end{cases}&
\end{flalign*}
        \end{minipage}
         \vspace{-1mm}
                \\
        \begin{minipage}[c]{\rightscalefactor\textwidth}
        \footnotesize
        \begin{flalign*}
&\pi^{(2)}
    \begin{cases}
    \includegraphics[width=0.95\textwidth]{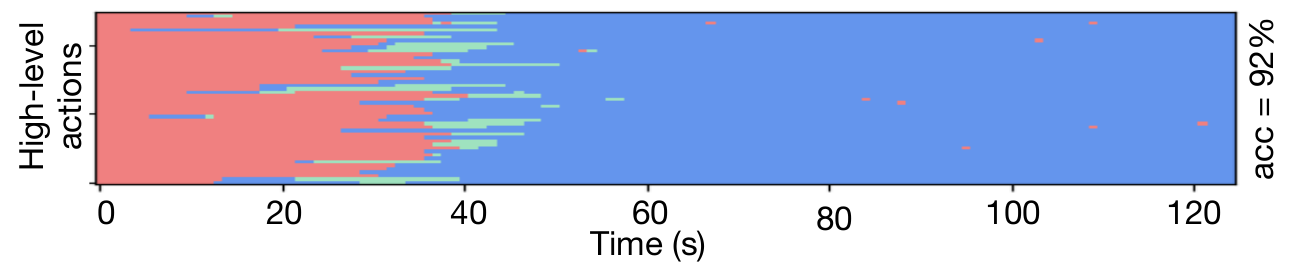}\\[-2mm]
       \;\;\;\;\;\;\guard_{\ACC,\CON}:= 
     {\stx{flp}(\stx{lgs}(v\!-\!v_{\max},0.5, 0.8))}   \\
         \;\;\;\;\;\;\guard_{\CON,\DEC}:=
   {\stx{flp}(\stx{lgs}(\dns, 1.6, -0.09)) }
    \end{cases}&
\end{flalign*}
        \end{minipage}
        \vspace{-1mm}
                \\
        \begin{minipage}[c]{\rightscalefactor\textwidth}
        \footnotesize
\begin{flalign*}
&
\pi^{(3)}
    \begin{cases}
     \includegraphics[width=0.95\textwidth]{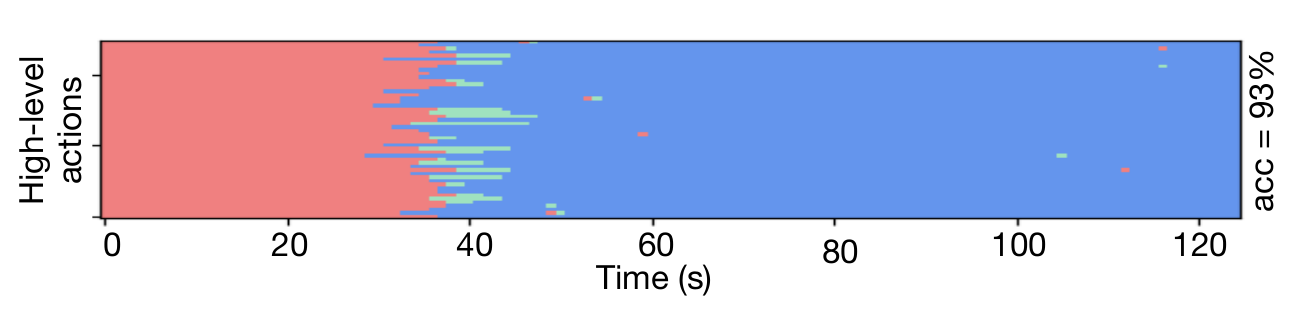}\\[-2mm]
       \;\;\;\;\;\;\guard_{\ACC,\CON}:= 
     {\stx{flp}(\stx{lgs}(v\!-\!v_{\max},0.08, 1.0))}   \\
          \;\;\;\;\;\;\guard_{\CON,\DEC}:=
   {\stx{flp}(\stx{lgs}(}
  {\stx{distTrv}(v,a_{\min})-\dns, 26.3, 0.07)) }
    \end{cases}&
\end{flalign*}
        \end{minipage}
        \vspace{-1mm}
\\
        \begin{minipage}[c]{\rightscalefactor\textwidth}
        \footnotesize
        \begin{flalign*}
&
\pi^{(5)} 
    \begin{cases}
    \includegraphics[width=0.95\textwidth]{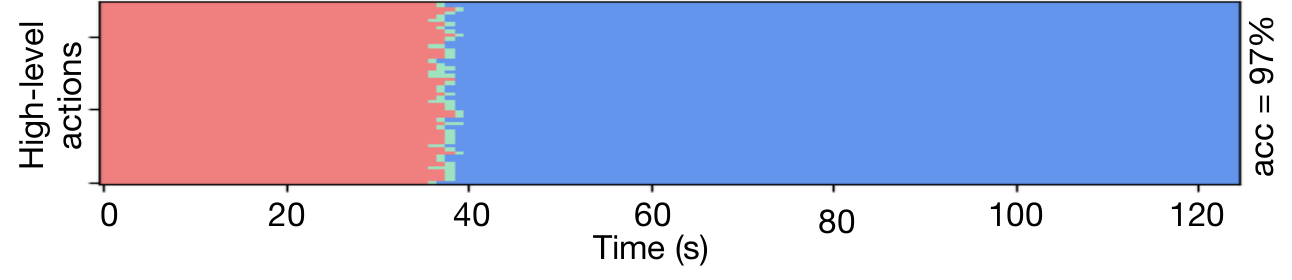}\\[-2mm]
     \;\;\;\;\;\;\guard_{\ACC,\CON}:= 
     {\stx{flp}(\stx{lgs}(v\!-\!v_{\max},-0.4, 1.3))}   \\
          \;\;\;\;\;\;\guard_{\CON,\DEC}:=
   {\stx{flp}(\stx{lgs}(}
  {\stx{distTrv}(v,a_{\min})-\dns, 2.8, 0.8)) }
    \end{cases}&
\end{flalign*}
        \end{minipage}

        \caption{
        Best candidate programs found at each iteration and the corresponding action sequence samples. The ground-truth sequence is shown at the top. 
       Only $\guard_{\ACC,\CON}$ and  $\guard_{\CON,\DEC}$ from each policy are shown due to space constraints. 
      } 
      \vspace{-3mm}
        \label{fig:em}
    \end{figure}

\autoref{fig:em} displays the results of four non-consecutive iterations of the EM loop for the Stop-Sign example. Each iteration presents $\pi^{(i)}$, the candidate policy synthesized during the M step, and its accuracy, alongside the corresponding action sequences sampled using the policy.
%
Due to space constraints, we only show the conditions $\guard_{\ACC,\CON}$ and $\guard_{\CON,\DEC}$ for each candidate policy. These conditions represent transitions from acceleration to constant velocity, and from constant velocity to deceleration, respectively.
Each sampled action sequence is depicted as a color-coded horizontal line. Specifically, $\ACC$ time-steps are shown in red, $\CON$ time-steps are shown in cyan, and $\DEC$ time-steps are shown in blue.
Because the candidate policies are probabilistic, the sequences from the same policy exhibit random variations. However, these variations diminish as the EM loop progresses and confidence in the synthesized policy increases.

%
%
 
%
The EM loop is initiated with a trivial ASP, $\pi^{(0)}$, employing $\stx{flp}(0.1)$ for all transition conditions.    
The algorithm then alternates between E and M Steps, progressively refining the candidate policy. This refinement continues until the desired level of accuracy is achieved in $\pi^{(5)}$, which is returned as the final solution for the Stop-Sign task.

In $\pi^{(5)}$, the transition condition $\guard_{\ACC,\CON}$ specifies that the vehicle should switch from acceleration to constant velocity, as its velocity nears the recommended maximum velocity, $v_{\max}$. 
Likewise, $\guard_{\CON,\DEC}$ indicates that the vehicle should transition from constant speed to deceleration when the estimated stop distance based on the current speed ($\stx{distTrv}$), approaches the remaining distance to the next stop sign ($\dns$).
These probabilistic expressions intuitively encapsulate the human demonstrators' intent for this specific task and are amenable to fine-tuning~\cite{srtr}.

%


Synthesizing ASPs such as $\pi^{(5)}$, is challenging due to their intricate structures with multiple functions, variables, and numerical constants. In the subsequent section, we will present more complex examples featuring multiple disjuncts and conjuncts drawn from our benchmarks.

\section{Experimental Evaluations}
\label{sec:eval}


We evaluated our approach using five standard imitation learning tasks from two challenging environments.

\noindent \textit{\textbf{Autonomous Vehicle Environment}}:
We utilize the open-source simulation environment \textsc{highway-env}~\cite{Eleurent2021}, where the demonstrator controls the acceleration and steering of a vehicle moving on a straight multi-lane highway. \newfinal{Policies have access to the position and velocity of the controlled vehicle and vehicles in adjacent lanes, and may select from action labels such as $\ha{accelerate}$, $\ha{decelerate}$, $\ha{turn\_left}$, and $\ha{turn\_right}$.}
We consider the following three tasks within this environment.
%
The {\bf Stop Sign (SS)} task is the motivating example described in previous sections. \old{and involves the demonstrator controlling the vehicle's acceleration in order to move to and stop at target locations.}
\old{
The Pass Traffic (PT) task is more complex and involves ongoing traffic. In this task, the agent's goal is to switch lanes quickly and safely to pass slower traffic.}
\new{The
\textbf{Pass Traffic (PT)} task requires
appropriately selecting and switching lanes in the highway traffic.
}
 %
 %
In the \textbf{Merge (MG)} task, the vehicle begins in the leftmost lane and must merge into the rightmost lane without colliding with traffic.

\noindent  \textit{\textbf{Robotic Arm Environment}}:
We also use the open-source \textsc{panda-gym} framework~\cite{gallouedec2021pandagym} that simulates a multi-joint robotic arm. Within this environment, the demonstrator manages both the end effector's movement in three-dimensional space and the signal to control its grip. 
\newfinal{Policies have access to the arm, object, and target locations, in three dimensions, and may select from action labels such as moving to an object, moving to the target, grasping an object, and lifting the arm.}
%
We evaluate our approach using two tasks defined in this environment. The first is the {\bf Pick and Place (PP)} task, which requires the arm to grab an object and then move it to a designated location \new{in an uncluttered environment}. 
The second is the {\bf Stack (ST)} task, which is more complex and requires stacking two boxes on top of each other in the correct order.

\noindent \new{ \textit{\textbf{Ground-truth Demonstrations}}:}
For each of the above five tasks, we generated a set of \new{$20$ to $30$} demonstrations \old{(\ie{} state and observation trajectories)} using a ground-truth policy that we implemented manually.
%
%
\new{
The demonstrations for each task were equally divided into a training set
and a test set. \autoref{fig:demos} presented the training set for the SS task.}
In addition to the \textit{transition noise} caused by the probabilistic nature of the ground-truth policies, the generated trajectories were also perturbed with \old{additional}\new{additive} Gaussian \textit{actuation noise}.
\old{Presence of actuation noise makes the policy-inference task more realistic, as such noise is commonly encountered when obtaining demonstrations in the real world. }
%
\old{
The demonstrations for each task were divided into a training set and a test set. Figure~2 presented the training set for the SS task.}

\noindent \new{ \textit{\textbf{Human-generated Demonstrations}}:}
\new{
In order to evaluate the baselines' ability to learn tasks under more realistic conditions, we also created datasets of human-generated demonstrations for PT and MG tasks using a dual-joystick controller for vehicle steering and acceleration. We refer to these datasets as \textbf{(PT-H)} and \textbf{(MG-H)}. These demonstrations have actuation noise similar to that of the ground truth demonstrations, and they exhibit higher transition noise due to the inherent inaccuracies of human behavior.
}

\subsection{Baselines}
We compare the performance of \algname{} against \old{five}\new{six} baselines, including \old{three}\new{four} state-of-the-art IL approaches.

    \noindent
    \textit{\textbf{Greedy}}: 
    Our first baseline represents a straightforward solution to the problem of imitation learning from unlabeled trajectories, where each time step is greedily labeled with the action whose likelihood is highest according to the observation model. The same policy synthesizer as \algname's M~step is then applied on these greedily labeled trajectories. 

    \noindent
    \textit{\textbf{OneShot}}: The next baseline synthesizes a policy from labels sampled using a particle filter and the  $\pi^{(0)}$ policy. This method can be viewed as the first iteration of \algname{}, except it does not employ incremental synthesis, and instead directly searches over a larger program space in one shot.


    %


    \noindent \textit{\textbf{LDIPS}}: Our third baseline is LDIPS~\cite{ldips}, a state-of-the-art PIL approach which synthesizes a policy via sketch enumeration and a reduction to satisfiability modulo theories (SMT). 
    Unlike \algname{}, LDIPS does not account for noise in the demonstrations and generates only deterministic policies. Furthermore, LDIPS requires action labels for synthesizing a policy. For this, we use the same action labels given to the OneShot baseline.

%


    

    \noindent
    \textit{\textbf{BC/BC+}}: 
    The next baseline (BC) implements Behavior Cloning~\cite{9117169}, a well-established imitation learning technique. \old{We trained an LSTM with 64 hidden units to predict the next observation conditioned on all previous states and observations.}\new{We trained a fully-connected feed-forward neural network with $3$ layers to predict the next observation conditioned on the previous state.}
    BC+ extends the BC baseline by incorporating access to action labels and the observation model. In BC+, rather than predicting observations directly, the \old{LSTM}\new{network} outputs a distribution over the available action labels. The observations are then predicted by performing a weighted sum based on the output of the observation model for each action label. 

    %

     
    \old{\textit{\textbf{GAIL}}:
Our final baseline is GAIL, another state-of-the-art method for imitation learning from expert trajectories. GAIL employs both a generator network and a discriminator network, using the principles of Generative Adversarial Networks (GANs).}
\noindent
\new{\textit{\textbf{GAIfO}}: Our next baseline is GAIfO~\cite{torabi2018generative}, a state-of-the-art method specifically designed for imitation learning from unlabeled demonstrations, using the principles of Generative Adversarial Networks (GANs)~\cite{goodfellow2014generative}.}

\old{Details of our setup for GAIL and hyper-parameters used for training are available in \textsc{Plunder}'s online repository.}

\noindent
\new{\textit{\textbf{Behavior Transformers}}: Our final baseline is Behavior Transformers (BeT)~\cite{shafiullah2022behavior}, a recent method that employs transformer-based sequence models  for imitation learning from a given set of continuous observation and action pairs. BeT is capable of multi-modal learning and provides support for both  Markovian and non-Markovian settings. } 

\begin{figure}[t]
\centering
\includegraphics[width=0.46\textwidth]{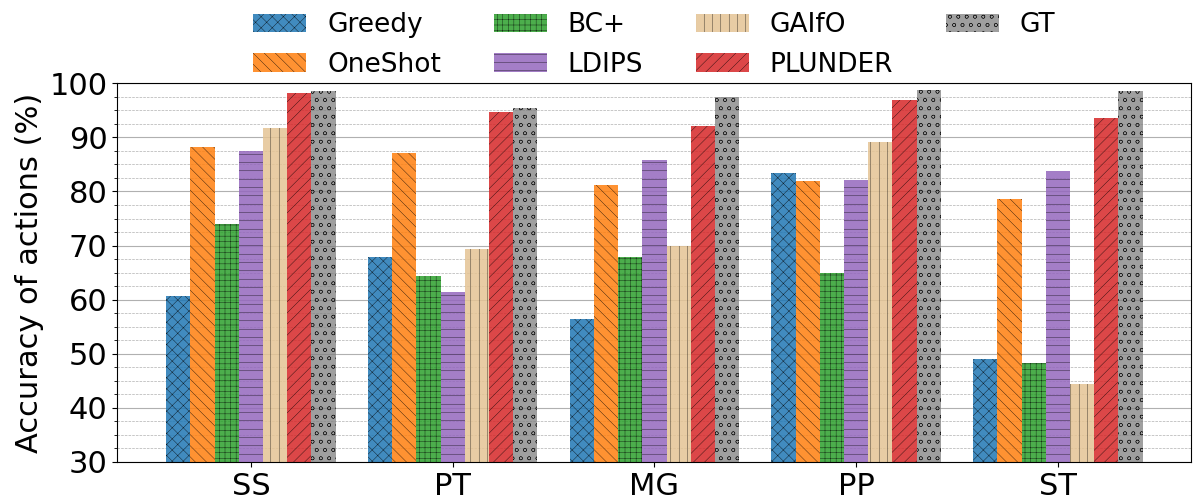}
    \caption{Accuracy of Action Labels 
    }
    \vspace{-2mm}
    \label{fig:action_accuracy}
\end{figure}
\begin{figure}[t] 
  \centering
\includegraphics[width=0.4\textwidth]{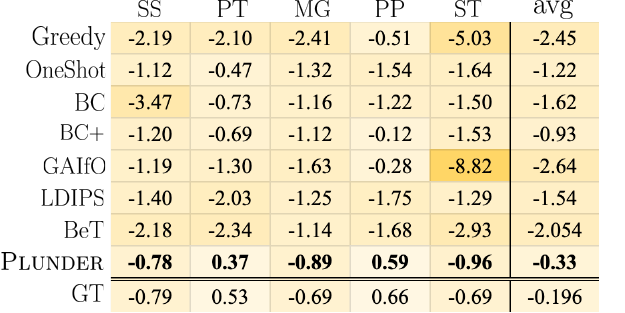}
    \caption{Log-Likelihood of Observations 
    }
     \vspace{-4mm}
    \label{fig:llh_table}
\end{figure}

\subsection{Alignment with Demonstrations} 
We evaluate the ability of \algname{} and the baselines to learn a policy that closely mimics the \new{ground-truth} demonstrations in the training set. 
%
\autoref{fig:action_accuracy} presents the accuracy of action labels each learned policy produces on the test trajectories. BC \new{and BeT} \old{is}\new{are} excluded because \old{it does}\new{they do} not produce action labels. \new{We also include results from the ground-truth policy (GT).}

The results reveal that \algname\ consistently outperforms all other baselines, achieving an average accuracy of $95\%$\new{, which is only $2.7\%$ below that of the ground truth policy}.  \old{We observe that the}\new{The} Greedy approach performs particularly poorly because of its tendency to over-fit to noise. The results confirm that utilizing a particle filter for inferring action labels, as implemented in the OneShot baseline, yields better performance compared to the na\"ive Greedy approach. 
We also find that the performance of LDIPS is significantly lower than that of \algname, confirming that deterministic policies are insufficient for capturing uncertainties present in the given demonstrations.

We also report the log-likelihood of observations generated by each policy in 
\autoref{fig:llh_table}. 
Again, across all evaluated tasks, \algname\ exhibits superior performance, yielding the highest log-likelihood. 

\begin{figure}[t]
\centering
\hspace{-4mm}
\begin{minipage}{0.23\textwidth}
        \includegraphics[width=1.07\linewidth]{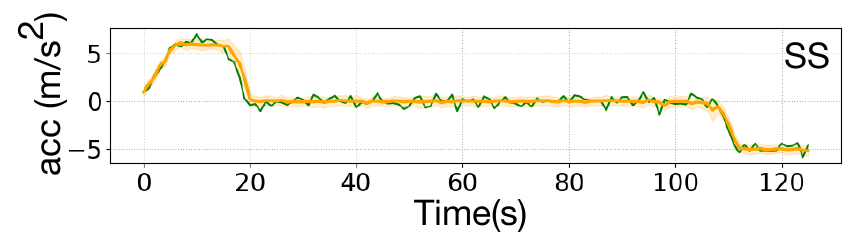}
        \end{minipage}
         \hspace{1mm}
        \begin{minipage}{0.23\textwidth}
         \includegraphics[width=1.07\linewidth]{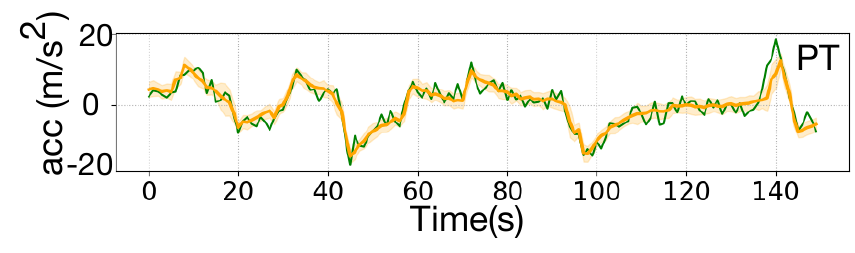}
         \end{minipage}
         \\[1mm]
         \hspace{-4mm}
         \begin{minipage}{0.23\textwidth}
        \includegraphics[width=1.07\linewidth]{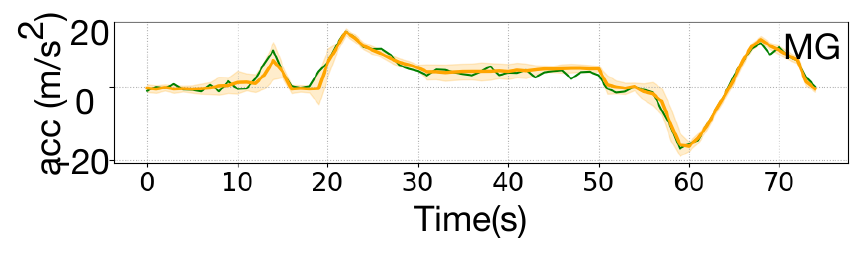}
        \end{minipage}
         \hspace{1mm}
        \begin{minipage}{0.23\textwidth}
         \includegraphics[width=1.07\linewidth]{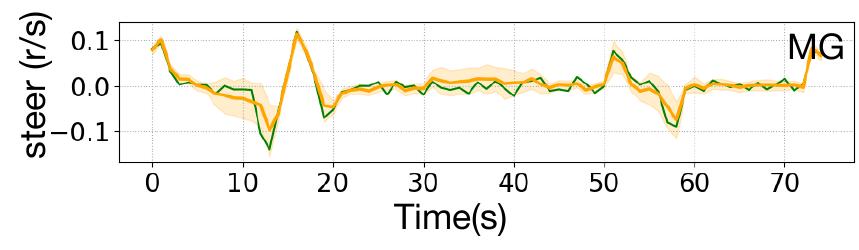}
         \end{minipage}
         \\[1mm]
         \hspace{-4mm}
          \begin{minipage}{0.23\textwidth}
        \includegraphics[width=1.07\linewidth]{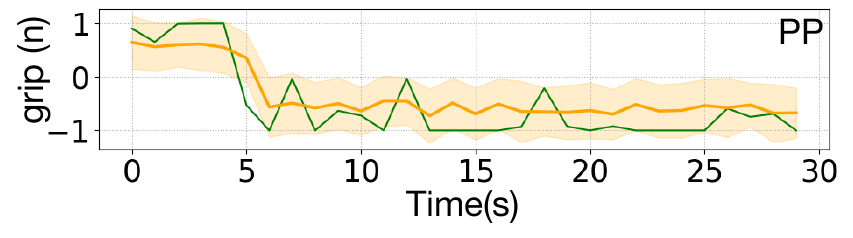}
        \end{minipage}
         \hspace{1mm}
        \begin{minipage}{0.23\textwidth}
         \includegraphics[width=1.07\linewidth]{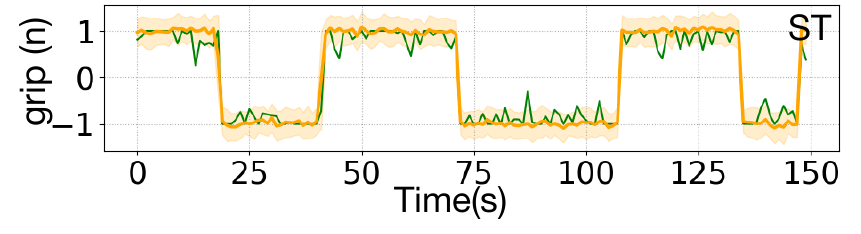}
         \end{minipage}
    \caption{Observation trajectories from the ground-truth policy (green line) and the \algname{} policy (orange line).}
    \label{fig:obs-plots}
    \vspace{-5mm}
\end{figure}

\autoref{fig:obs-plots} visualizes samples of the ground-truth  trajectories (in green) and the corresponding trajectories generated by the \algname\ policy (in orange). 
The trajectories synthesized by \algname{} closely align with the ground-truth across all benchmarks. 
Due to space limitations, we include a subset of observations from each task; however, we witnessed similar behaviors across all observations.


\begin{figure*}[t]
\centering
\begin{subfigure}[t]{0.53\textwidth}
    \includegraphics[width=\linewidth]{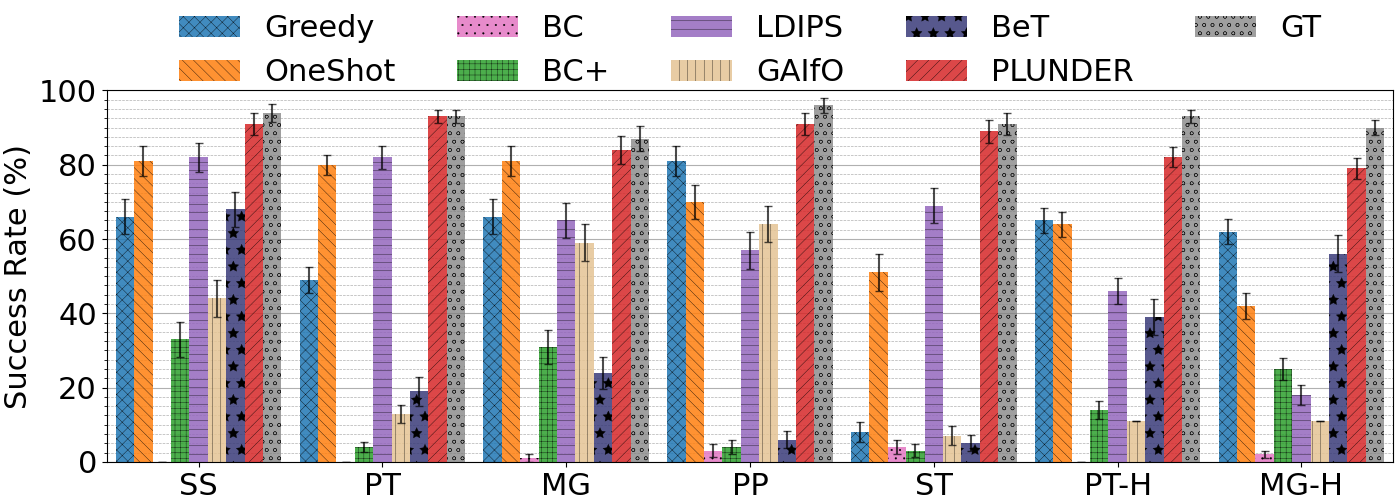}
    \caption{Success Rate}
    \label{subfig:success}
\end{subfigure}
\hfill
\begin{subfigure}[t]{0.22\textwidth}
    \centering
    \hspace{-2mm}\includegraphics[width=1.03\linewidth]{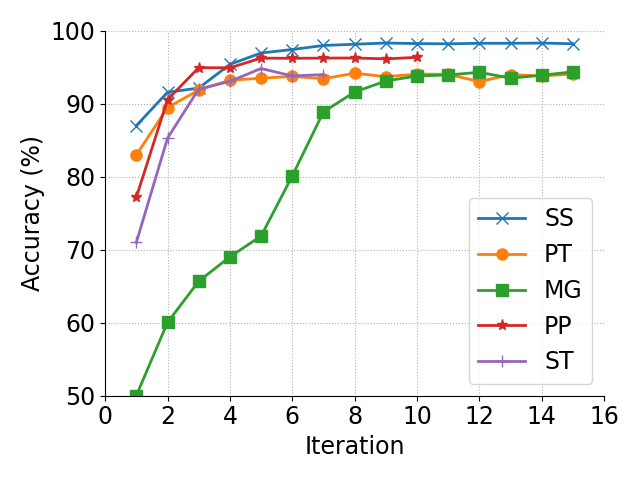}
    \caption{Convergence}
    \label{subfig:convergence}
\end{subfigure}
\hfill
\begin{subfigure}[t]{0.23\textwidth}
    \centering
    \hspace{-2mm}\includegraphics[width=1.03\textwidth]{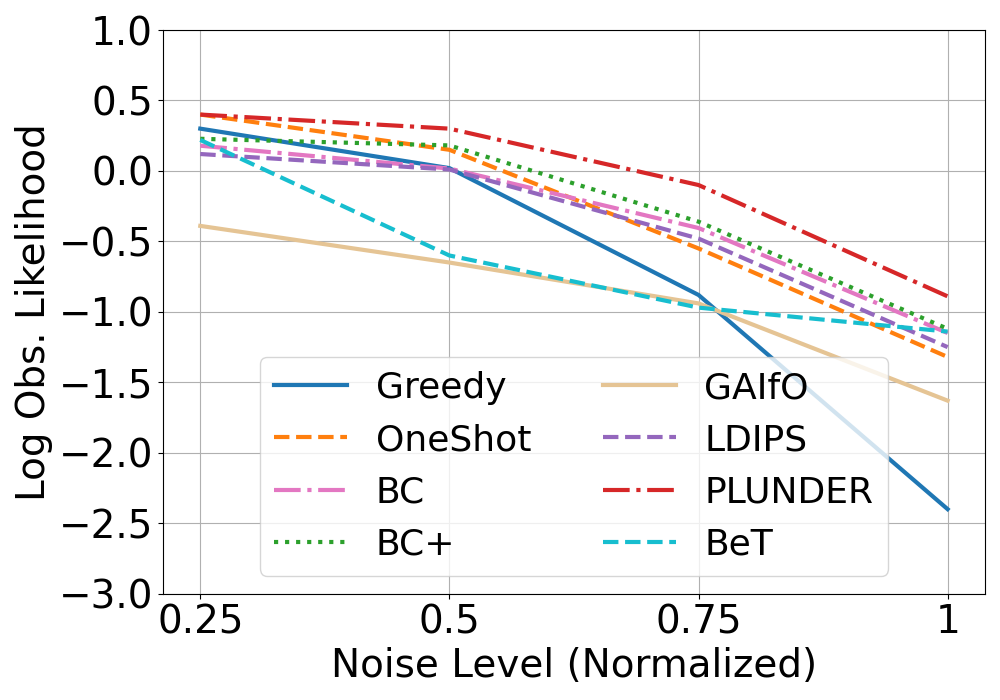}
    \caption{Noise Tolerance}
    \label{subfig:noise}
\end{subfigure}        
    \caption{Empirical results: (a) the success rates of policies learned across all baseline approaches, (b) the progression of the EM loop until convergence, and (c) the impact of noise on the learning performance of each baseline approach. }
    \vspace{-3mm}
\end{figure*}

\subsection{Task Completion Rate}
We next evaluate the success rate of the learned policies
using \algname{} and other baselines. Each learned policy was executed 100 times on randomly initialized environments and we report the percentage of executions where the task was successfully completed.

\old{The results are presented in Figure 8.
}
\new{The success rates of each baseline, with 1 SD error bars, are presented in \autoref{subfig:success}.
}
\algname{} achieves an average success rate of 90\%, the highest among all baselines. In contrast, the BC and BC+ approaches perform particularly poorly. We attribute this to the compounding errors effect and the inherent inability of BC methods to generalize from a small training set.
\new{
Notably, the gap between \algname{} and all other baselines is larger in PT-H and MG-H with more realistic demonstration datasets. 
}

\subsection{Convergence}
Our experiments support our hypothesis that the accuracy of the synthesized policies increases with each iteration of the EM loop. 
\autoref{subfig:convergence} illustrates the performance of the synthesized policy after each EM loop iterations. 
\algname{} converges to a policy across all tasks in fewer than 10 iterations. 

\subsection{Impact of Noise}
We evaluate the performance of each baseline as we vary the amount of actuation noise in the training data for the {MG} task. 
The \old{accuracies and} observation log likelihoods of the trajectories produced by all methods are depicted in \autoref{subfig:noise}. 
All approaches show a decrease in performance as the noise increases; however, \algname{} consistently surpasses other baselines, indicating its superior robustness against noise. 
%
We also note that OneShot is more robust against noise compared to Greedy. This underscores the significance of employing a particle filter, even when backed by a simplistic prior. 
Lastly, the results indicate that {BC+} has an advantage over {BC}, hinting that having access to the observation model provides useful inductive bias to overcome the noise in the data.

\subsection{Analysis }
\old{We found that in all tasks, the synthesized 
programs contain non-trivial conditions that require multiple incremental synthesis steps to be discovered.}
\begin{table}[t]
\centering
\footnotesize
\new{
\begin{tabular}{lcccccccc}
\hline
Benchmark & SS & PT & MG & PP & ST  & PT-H & MG-H \\ \hline
AST Size  & 25  & 96  & 46  & 22  & 55  & 196 & 98   \\ \hline
\end{tabular}
}
\caption{\new{AST Size of Synthesized Policies}}
\label{tab:ast}
\vspace{-5mm}
\end{table}
\new{\autoref{tab:ast} shows the number of nodes in the AST (Abstract Syntax Tree) of the synthesized policies for each task. 
We found that in all tasks, the synthesized 
policies contain non-trivial conditions that require multiple incremental synthesis steps to be discovered.
}
For instance, the following transition condition was synthesized in the {PT} task:
$$
 \begin{array}{ll}
\guard_{\ha{FASTER},\ha{LANE\_LEFT}} & = \stx{flp}(\stx{lgs}(x - f_x, -30.62, 1.78))  \\
&  \wedge\ \stx{flp}(\stx{lgs}(f_x - l_x, -4.92, -6.53)) \\ 
&  \wedge\ \stx{flp}(\stx{lgs}(r_x - l_x, 2.95, -1.67))
 \end{array}
$$

The expression above accurately reflects the behaviors observed in the demonstrations. It states:
``If the vehicle's position \((x)\) is approaching the position of the vehicle
in front \((f_x)\), and the position of the vehicle to the left \((l_x)\) is
farther away than both the vehicle in front and the position of the vehicle to
the right \((r_x)\), then switch to the left lane''.
Such expressions are amenable to \newfinal{refinement }and formal verification, which is a significant advantage of \algname{} over methods such as BC, BC+, GAIfO, and BeT.
Note that expressions of this complexity exceed the capabilities of na\"ive enumeration techniques, which would need to optimize real-valued parameters across more than a billion sketches to identify such a policy. Such enumeration would not be feasible within an EM loop.

\section{Conclusions\new{, Limitations,} and Future Work} 
\label{sec:conclusion}
\vspace{0.2em}
{We presented \textsc{Plunder}, an algorithm for inferring probabilistic programs from noisy and unlabeled demonstrations. A reusable implementation of this algorithm, as well as all empirical evaluation results, are made available online.}
\new{
Our approach encounters the typical limitations associated with syntax-based enumerative program synthesis algorithms, namely, the exponential growth in the search space as the size of the DSL and the complexity of tasks increase. In \algname{}, we  achieved tractable synthesis through careful design of the DSL and the application of a series of heuristics. \newfinal{Additionally, the EM framework is generally susceptible to local optima. In \algname{}, we reduce the risk of encountering such local solutions by performing multiple runs of the synthesizer from random initial seeds, and by incorporating simple atomic policies outside the syntactic neighborhood of the current policy.} For future work, we plan to devise a more general solution by leveraging recent advancements in program synthesis algorithms, such as neural-guided program search~\cite{chaudhuri2021neurosymbolic} and those utilizing Large Language Models~\cite{patton2023program}\newfinal{, and apply these solutions to real-world robot data.}
}
We also aim to minimize the required user input by optimizing the observation model concurrently\new{, and introduce an automated parameter adjustment mechanism for the hyper-parameter $\lambda$}.
\old{Additionally, we plan to explore other advanced synthesis techniques, such as those utilizing Large Language Models, within the context of the EM loop to further enhance the usability of this approach.}

\section{Acknowledgement}
\newfinal{
We would like to thank our anonymous reviewers for their helpful and insightful feedback.
This work is partially supported by the National Science Foundation (CAREER-2046955, CCF-2319471,
CCF-1762299, CCF-1918889, CNS-1908304, CCF-1901376, CNS-2120696, CCF-2210831). Any opinions, findings, and conclusions expressed in this material are those of
the authors and do not necessarily reflect the views of the sponsors.
}

\vspace{-2mm}
\bibliography{main} 
\bibliographystyle{ieeetr}

\end{document}